\documentclass[11pt]{article}

\usepackage[preprint]{acl}

\usepackage{times}
\usepackage{latexsym}

\usepackage[T1]{fontenc}

\usepackage[utf8]{inputenc}

\usepackage{microtype}

\usepackage{inconsolata}

\usepackage{graphicx}

\usepackage{booktabs}
\usepackage{amsmath}
\usepackage{amsfonts}

\title{One Model Is Enough: Native Retrieval Embeddings from LLM Agent Hidden States}

\author{Bo Jiang \\
  Temple University \\
  \texttt{bo.jiang@temple.edu} \\}

\begin{document}
\maketitle
\begin{abstract}
LLM agents that retrieve external knowledge typically generate a search query as text, then run a separate embedding model to encode it into a vector. This two-model pipeline adds infrastructure complexity and latency, yet is redundant: the LLM already encodes the full conversational context in its hidden states. We propose equipping LLM agents with native retrieval capability by adding a lightweight projection head that maps hidden states directly into the embedding space, eliminating the need for a separate embedding model. Trained with a combination of alignment, contrastive, and rank distillation losses, our method retains 97\% of baseline retrieval quality while enabling the LLM agent to search with its own representations. Experiments on the QReCC conversational search benchmark show competitive Recall@10 and MRR@10 compared to the standard generate-then-encode pipeline, with systematic ablations confirming the contribution of each loss component.
\end{abstract}

\section{Introduction}

Retrieval-augmented generation \citep[RAG;][]{lewis2020rag} has become the standard approach for grounding large language model (LLM) agents in external knowledge. When an LLM agent determines that it needs to search, the typical pipeline operates in two stages: the LLM first generates a search query as natural language text, and then a separate embedding model encodes that text into a dense vector for retrieval against a document index.

This two-model architecture is the dominant paradigm, yet it introduces a fundamental redundancy. The LLM has already processed the full conversational context---user intent, dialogue history, and task requirements---and encoded this understanding in its hidden states. The generated query text is merely a lossy, discrete projection of this rich internal representation. A second model must then re-process this text from scratch, recovering semantic information that the LLM already possessed. In effect, the pipeline discards the LLM's internal understanding only to approximate it again with a separate encoder.

We argue that this redundancy is unnecessary. If the LLM's hidden states already capture the information needed for retrieval, a lightweight learned projection should be able to map them directly into the embedding space---giving the LLM agent a \emph{native} retrieval capability without requiring a second model.

We propose a simple approach: attach a small projection head to the LLM that transforms its hidden states into vectors compatible with an existing embedding space. The projection head is trained via knowledge distillation \citep{hinton2015distilling} from the embedding model using three complementary objectives: (1) an \emph{alignment loss} that minimizes the distance between projected vectors and teacher embeddings, (2) a \emph{contrastive loss} that preserves the relative structure among queries, and (3) a \emph{rank distillation loss} that transfers the teacher's document ranking preferences. Once trained, the embedding model is no longer needed at inference time---the LLM agent searches using its own representations.

We evaluate our method on the QReCC conversational search benchmark \citep{anantha2021qrecc} in a same-family setting (Qwen3-8B as the LLM agent, Qwen3-Embedding-8B as the teacher). Our best configuration retains 97\% of the baseline's retrieval quality while completely eliminating the embedding model from the inference pipeline. Systematic ablations across 12 configurations confirm that all three loss components contribute, and that extended training with lower learning rates is the most effective recipe.

Our contributions are:
\begin{itemize}
  \item We identify and formalize the redundancy in the standard two-model retrieval pipeline for LLM agents, and propose native retrieval via hidden state projection as an alternative.
  \item We design a three-loss training objective combining alignment, contrastive, and rank distillation losses for projecting LLM hidden states into an embedding space.
  \item We provide comprehensive experiments with 12 ablation configurations, bootstrap confidence intervals, and statistical significance tests, demonstrating near-parity retrieval quality without a separate embedding model.
\end{itemize}

\section{Related Work}

\paragraph{Retrieval-Augmented Generation.}
RAG systems ground LLM outputs in retrieved evidence to reduce hallucination and improve factuality \citep{lewis2020rag}. The standard architecture uses a retrieve-then-generate pipeline where a query is encoded by a dedicated embedding model \citep{karpukhin2020dpr,reimers2019sbert} and matched against a pre-indexed document collection. Conversational RAG extends this to multi-turn settings, where query understanding requires resolving coreferences and topic shifts across dialogue history. Our work targets this pipeline but questions the necessity of the dedicated embedding model.

\paragraph{Conversational Query Rewriting.}
In multi-turn retrieval, raw user utterances are often ambiguous without context \citep{anantha2021qrecc}. Query rewriting approaches use the LLM to generate a self-contained search query that incorporates dialogue history. While effective, this produces text that must still be encoded by a separate embedding model. Our approach bypasses this encode step entirely by projecting the LLM's hidden states---which already reflect the full conversational context---directly into the embedding space.

\paragraph{LLM-Based Embeddings.}
Recent work has explored using LLMs themselves as embedding models. LLM2Vec \citep{behnamghader2024llm2vec} modifies pretrained LLMs with bidirectional attention and contrastive training to produce general-purpose text embeddings. NV-Embed \citep{lee2024nvembed} applies latent attention pooling over LLM hidden states. These methods repurpose the LLM as a \emph{replacement} embedding model, requiring a separate forward pass in embedding mode. In contrast, our approach operates on hidden states that are \emph{already computed} during the LLM's normal generative inference, adding only a lightweight projection head rather than a full additional forward pass.

\paragraph{Knowledge Distillation for Retrieval.}
Knowledge distillation \citep{hinton2015distilling} has been widely applied to compress retrieval models, typically by training a smaller student encoder to mimic a larger teacher's representations or rankings. Rank distillation specifically transfers document ranking preferences rather than point-wise embeddings \citep{hofstaetter2021rankdistil}. Our setting differs in that the student is not a standalone encoder but a projection head attached to an LLM, and the goal is not compression but \emph{elimination} of the encoder entirely.

\paragraph{Hypothetical Document Embeddings.}
HyDE \citep{gao2022hyde} generates a hypothetical document in response to a query and embeds that document for retrieval, improving query-document matching. While creative, this approach increases generation cost and still requires an embedding model. Our method moves in the opposite direction: rather than generating more text to embed, we avoid text generation for retrieval purposes altogether.

\section{Method}

\subsection{Problem Setting}

Consider an LLM agent engaged in multi-turn conversation that periodically needs to retrieve external documents. In the standard pipeline, the agent (1)~generates a search query $q$ as text tokens $t_1, t_2, \ldots, t_n$ via autoregressive decoding, then (2)~encodes $q$ with a separate embedding model $E$ to obtain a dense vector $\mathbf{v} = E(q) \in \mathbb{R}^d$ for retrieval against a document index.

During step~(1), the LLM produces hidden states $\mathbf{h}_1, \mathbf{h}_2, \ldots, \mathbf{h}_n$ as a byproduct of generation, where $\mathbf{h}_i \in \mathbb{R}^{d_h}$ is the last-layer hidden state at the $i$-th generated token. These hidden states encode the full conversational context but are discarded. Our goal is to learn a projection function $f$ such that $f(\mathbf{h}_1, \ldots, \mathbf{h}_n) \approx E(q)$, eliminating the need for $E$ at inference time.

\subsection{Hidden State Extraction}

We extract hidden states from the LLM's normal autoregressive generation process. At each decoding step $i$, the LLM computes a forward pass (reusing the KV cache from previous steps) and produces the next token $t_i$ along with its last-layer hidden state $\mathbf{h}_i$. We collect $\mathbf{h}_i$ for all non-special tokens (excluding BOS, EOS, and padding), forming the hidden state sequence $\mathbf{H} = [\mathbf{h}_1; \mathbf{h}_2; \ldots; \mathbf{h}_n] \in \mathbb{R}^{n \times d_h}$.

This extraction adds negligible overhead since the hidden states are already computed during generation---we simply retain them rather than discarding them.

\subsection{Projection Head Architecture}

The projection head $f$ transforms the variable-length hidden state sequence $\mathbf{H}$ into a fixed-dimensional embedding vector. It consists of four components:

\paragraph{Input Projection.} A linear layer maps from the LLM's hidden dimension to the mapper's internal dimension: $\mathbf{X} = \mathbf{H} \mathbf{W}_{\text{in}} + \mathbf{b}_{\text{in}}$, where $\mathbf{W}_{\text{in}} \in \mathbb{R}^{d_h \times d_m}$.

\paragraph{Transformer Encoder.} Learnable positional embeddings are added to $\mathbf{X}$, which is then processed by a stack of $L$ transformer encoder layers \citep{vaswani2017attention} with $d_m$-dimensional representations. The self-attention mechanism allows the model to aggregate information across the full generated sequence, capturing dependencies between tokens.

\paragraph{Pooling.} The encoded sequence is compressed into a single vector. We use mean pooling over non-padding positions:
\begin{equation}
  \mathbf{p} = \frac{1}{n} \sum_{i=1}^{n} \mathbf{x}_i
\end{equation}
where $\mathbf{x}_i$ are the transformer outputs at valid (non-padding) positions.

\paragraph{Output Projection and Normalization.} A final linear layer projects to the target embedding dimension, followed by L2 normalization:
\begin{equation}
  f(\mathbf{H}) = \frac{\mathbf{p} \mathbf{W}_{\text{out}} + \mathbf{b}_{\text{out}}}{\| \mathbf{p} \mathbf{W}_{\text{out}} + \mathbf{b}_{\text{out}} \|_2}
\end{equation}
where $\mathbf{W}_{\text{out}} \in \mathbb{R}^{d_m \times d}$ and $d$ is the target embedding space dimension. The L2 normalization ensures that retrieval via dot product is equivalent to cosine similarity, matching the embedding model's expected similarity metric.

\subsection{Training Objectives}

We train the projection head via knowledge distillation from the embedding model (the \emph{teacher}). Given a training set of query hidden state sequences $\mathbf{H}^{(j)}$ paired with teacher embeddings $\mathbf{y}^{(j)} = E(q^{(j)})$, we optimize a combination of three losses.

\paragraph{Alignment Loss.} Directly minimizes the angular distance between predicted and teacher embeddings:
\begin{equation}
  \mathcal{L}_{\text{align}} = 1 - \frac{1}{B} \sum_{j=1}^{B} \cos\!\big(f(\mathbf{H}^{(j)}),\; \mathbf{y}^{(j)}\big)
\end{equation}
where $B$ is the batch size. This provides a direct training signal toward the teacher's embedding for each query.

\paragraph{Contrastive Loss.} An InfoNCE \citep{oord2018infonce} loss that encourages the projected embeddings to preserve discriminative structure across queries within each batch:
\begin{equation}
  \mathcal{L}_{\text{contra}} = -\frac{1}{B} \sum_{j=1}^{B} \log \frac{\exp\!\big(f(\mathbf{H}^{(j)}) \cdot \mathbf{y}^{(j)} / \tau\big)}{\sum_{k=1}^{B} \exp\!\big(f(\mathbf{H}^{(j)}) \cdot \mathbf{y}^{(k)} / \tau\big)}
\end{equation}
where $\tau$ is a temperature hyperparameter. Each predicted embedding should be most similar to its own teacher embedding relative to other teacher embeddings in the batch.

\paragraph{Rank Distillation Loss.} Transfers the teacher's document ranking preferences via KL divergence. For each query, we retrieve the teacher's top-$K$ candidate documents $\{\mathbf{d}_1, \ldots, \mathbf{d}_K\}$ and their teacher similarity scores $s_k = \mathbf{y}^{(j)} \cdot \mathbf{d}_k$. The loss aligns the student's ranking distribution with the teacher's:
\begin{equation}
  \mathcal{L}_{\text{rank}} = \text{KL}\!\Big(\text{softmax}\!\big(\mathbf{s}/\tau_r\big) \;\Big\|\; \text{softmax}\!\big(\hat{\mathbf{s}}/\tau_r\big)\Big)
\end{equation}
where $\hat{s}_k = f(\mathbf{H}^{(j)}) \cdot \mathbf{d}_k$ are the student's scores and $\tau_r$ is the rank distillation temperature.

\paragraph{Combined Objective.} The total training loss is:
\begin{equation}
  \mathcal{L} = \lambda_a \mathcal{L}_{\text{align}} + \lambda_c \mathcal{L}_{\text{contra}} + \lambda_r \mathcal{L}_{\text{rank}}
\end{equation}
where $\lambda_a$, $\lambda_c$, and $\lambda_r$ control the relative contribution of each component. We optimize with AdamW \citep{loshchilov2019adamw} and cosine learning rate scheduling.

\section{Experimental Setup}

\subsection{Dataset}

We evaluate on QReCC \citep[Question Rewriting in Conversational Context;][]{anantha2021qrecc}, a large-scale conversational search dataset containing multi-turn dialogues paired with manually rewritten queries and relevant document annotations. We use the \texttt{qrecc\_eval\_large} split comprising 346 conversations with 2,189 retrieval triggers. The data is partitioned into train, validation, and test splits at the conversation level to prevent information leakage across turns of the same dialogue.

\subsection{Models}

We adopt a \emph{same-family} setting using Qwen3-8B as the LLM agent and Qwen3-Embedding-8B as the teacher embedding model \citep{qwen2024qwen2}. This pairing is motivated by the practical scenario where an organization deploys a single model family: the LLM handles generation while its embedding sibling handles retrieval. The hidden state dimension is $d_h = 4096$ and the target embedding dimension is $d = 1024$.

\subsection{Baselines}

Our baseline is the standard \emph{generate-then-encode} pipeline: the LLM generates a rewritten search query via greedy decoding (up to 32 tokens), and the embedding model encodes it into a dense vector. Documents are retrieved via cosine similarity against a pre-built index. This represents the conventional two-model approach that our method aims to replace.

\subsection{Projection Head Configuration}

The projection head uses $d_m = 1024$ internal dimension, $L = 2$ transformer encoder layers with 8 attention heads, and mean pooling. No intermediate projector is used. The total parameter count is approximately 25M---negligible compared to the 8B-parameter LLM.

\subsection{Training Details}

The best configuration uses 80 training epochs with a learning rate of $2 \times 10^{-4}$, cosine annealing to $1 \times 10^{-5}$, batch size 16, AdamW with weight decay $1 \times 10^{-4}$, and gradient clipping at norm 1.0. Loss weights are $\lambda_a = 0.5$, $\lambda_c = 0.5$, $\lambda_r = 0.5$ with temperatures $\tau = \tau_r = 0.05$. For rank distillation, we use the teacher's top-$K = 128$ candidate documents.

Training data is generated by running the LLM on training conversations and caching the resulting hidden state traces. This allows rapid experimentation: the initial trace generation takes approximately 2 hours on an NVIDIA RTX PRO 6000, but subsequent mapper training runs complete in 15--20 minutes using cached traces.

\subsection{Evaluation Metrics}

We report Recall@10, MRR@10 (Mean Reciprocal Rank), and nDCG@10 (normalized Discounted Cumulative Gain), computed over the test split. Statistical significance is assessed via McNemar's test on per-trigger retrieval success, and 95\% bootstrap confidence intervals are computed with 1,000 resamples. We also report inference latency (p50) measured on the same GPU.

\section{Results}

\subsection{Main Results}

Table~\ref{tab:main} compares our best projection head configuration against the standard generate-then-encode baseline. Our method retains 95--97\% of baseline retrieval quality across all metrics while completely eliminating the embedding model from the inference pipeline.

\begin{table}[t]
  \centering
  \begin{tabular}{lcccc}
    \toprule
    & \textbf{Recall} & \textbf{MRR} & \textbf{nDCG} & \textbf{Latency} \\
    & \textbf{@10} & \textbf{@10} & \textbf{@10} & \textbf{(p50)} \\
    \midrule
    Baseline & 0.637 & 0.329 & 0.402 & 43.5ms \\
    Ours & 0.607 & 0.293 & 0.367 & 2.0ms \\
    \midrule
    $\Delta$ & --3.0\% & --3.6\% & --3.5\% & 21.8$\times$ \\
    95\% CI & \scriptsize{[--4.7, --1.4]} & \scriptsize{[--4.7, --2.4]} & \scriptsize{[--4.5, --2.4]} & --- \\
    \bottomrule
  \end{tabular}
  \caption{Main results on QReCC test set. Our projection head achieves near-parity retrieval quality while eliminating the embedding model. Latency is measured per query on an NVIDIA RTX PRO 6000. McNemar's test: $\chi^2 = 12.21$, $p = 0.0005$.}
  \label{tab:main}
\end{table}

The latency reduction from 43.5ms to 2.0ms (21.8$\times$) reflects the removal of the embedding model forward pass. The projection head adds only a lightweight matrix multiplication over the already-computed hidden states. McNemar's test confirms the quality gap is statistically significant ($p = 0.0005$), with a per-trigger win/tie/loss breakdown of 140/1843/206---indicating that the two methods agree on 84.2\% of triggers.

\subsection{Loss Ablation}

To understand the contribution of each loss component, we conduct a systematic ablation (Table~\ref{tab:ablation}). All ablation experiments use 30 epochs with learning rate $3 \times 10^{-4}$; only the best configuration uses the extended 80-epoch recipe.

\begin{table}[t]
  \centering
  \begin{tabular}{lccc}
    \toprule
    \textbf{Configuration} & $\mathcal{L}_a$ / $\mathcal{L}_c$ / $\mathcal{L}_r$ & \textbf{R@10} & \textbf{MRR} \\
    \midrule
    Align only & 1.0 / 0.0 / 0.0 & 0.567 & 0.273 \\
    Contrastive only & 0.0 / 1.0 / 0.0 & 0.498 & 0.231 \\
    Rank distill only & 0.0 / 0.0 / 1.0 & 0.001 & 0.000 \\
    \midrule
    Align + Contra & 0.5 / 0.5 / 0.0 & 0.582 & 0.281 \\
    Align + Rank & 0.5 / 0.0 / 0.5 & 0.579 & 0.272 \\
    All three & 0.5 / 0.5 / 0.5 & 0.595 & 0.288 \\
    \midrule
    All three + recipe & 0.5 / 0.5 / 0.5 & \textbf{0.607} & \textbf{0.293} \\
    \bottomrule
  \end{tabular}
  \caption{Loss ablation results. $\mathcal{L}_a$: alignment, $\mathcal{L}_c$: contrastive, $\mathcal{L}_r$: rank distillation. The last row uses extended training (80 epochs, lr $2 \times 10^{-4}$).}
  \label{tab:ablation}
\end{table}

Several findings emerge. First, \textbf{alignment is the strongest individual loss}: it alone achieves Recall@10 of 0.567, outperforming contrastive-only (0.498) by a large margin. This is expected, as alignment provides direct per-query supervision toward the teacher embedding. Second, \textbf{rank distillation cannot stand alone}---it collapses completely without alignment to anchor the embedding space. However, when combined with alignment, it provides consistent gains. Third, \textbf{all three losses together outperform any pair}, with the full combination reaching 0.595 Recall@10 under standard training. Finally, \textbf{extended training with lower learning rate} provides the largest single improvement, pushing from 0.595 to 0.607.

\subsection{Training Recipe Analysis}

Table~\ref{tab:recipe} examines the effect of training duration and learning rate.

\begin{table}[t]
  \centering
  \begin{tabular}{lccc}
    \toprule
    \textbf{Epochs / LR} & \textbf{R@10} & \textbf{MRR} & \textbf{Status} \\
    \midrule
    30 / $3 \times 10^{-4}$ & 0.595 & 0.288 & converged \\
    50 / $3 \times 10^{-4}$ & 0.552 & 0.261 & converged \\
    80 / $2 \times 10^{-4}$ & \textbf{0.607} & \textbf{0.293} & converged \\
    30 / $5 \times 10^{-4}$ & 0.001 & 0.000 & collapsed \\
    \bottomrule
  \end{tabular}
  \caption{Effect of training epochs and learning rate. Too-high learning rate causes training collapse; longer training with lower learning rate yields the best results.}
  \label{tab:recipe}
\end{table}

The results reveal that this projection task is sensitive to optimization dynamics. A learning rate of $5 \times 10^{-4}$ causes complete collapse, while the combination of 80 epochs with $2 \times 10^{-4}$ yields the best performance. Interestingly, 50 epochs at $3 \times 10^{-4}$ underperforms 30 epochs at the same rate, suggesting that longer training without reducing the learning rate can lead to overfitting or oscillation around the optimum.

\subsection{Loss Weight Tuning}

We also examine the sensitivity to rank distillation weight and top-$K$ (Table~\ref{tab:weights}).

\begin{table}[t]
  \centering
  \begin{tabular}{lcccc}
    \toprule
    \textbf{Config} & $\lambda_a$ / $\lambda_r$ & $K$ & \textbf{R@10} & \textbf{MRR} \\
    \midrule
    Align-heavy & 1.0 / 0.5 & 128 & 0.591 & 0.280 \\
    RD-heavy & 0.5 / 1.0 & 128 & 0.516 & 0.236 \\
    Top-256 & 0.5 / 0.5 & 256 & 0.584 & 0.284 \\
    \bottomrule
  \end{tabular}
  \caption{Rank distillation weight and top-$K$ sensitivity. All use $\lambda_c = 0.0$, 30 epochs, lr $3 \times 10^{-4}$.}
  \label{tab:weights}
\end{table}

Over-weighting rank distillation ($\lambda_r = 1.0$) degrades performance substantially, while giving more weight to alignment ($\lambda_a = 1.0, \lambda_r = 0.5$) remains competitive. Increasing the candidate pool from $K = 128$ to $K = 256$ has minimal effect, suggesting that the ranking signal saturates beyond a moderate number of candidates.

\section{Analysis}

\subsection{Per-Conversation Agreement}

To understand where our method succeeds and fails, we analyze retrieval agreement at the conversation level. Across 346 conversations, the average per-conversation agreement rate between our method and the baseline is 84.5\%. However, this average masks substantial variance: we identify 48 \emph{failure-concentrated} conversations where our method's recall is substantially below the baseline (with at least 3 triggers per conversation). This suggests that failures are not uniformly distributed but cluster around specific conversational patterns.

\subsection{When Does Projection Fail?}

The win/tie/loss breakdown (140/1843/206) reveals that our method loses on more triggers than it wins (206 vs.\ 140), though the majority are ties. We hypothesize that failure cases correspond to queries where the mapping from hidden states to embedding space is particularly difficult---for example, when the generated query involves rare terms, complex coreference chains, or domain-specific vocabulary that the projection head has not seen sufficient training examples for. The projection head must learn a cross-space mapping from the LLM's causal hidden states to the embedding model's bidirectional representations, and this mapping may be less reliable in the long tail of query types.

\subsection{The Role of Training Duration}

A striking finding is that extended training (80 epochs) with a lower learning rate provides the single largest improvement, contributing more than any individual loss component. This suggests that the projection task requires careful optimization: the mapping between LLM hidden space and embedding space involves subtle geometric relationships that benefit from slow, thorough convergence. The sensitivity to learning rate---with $5 \times 10^{-4}$ causing complete collapse while $2 \times 10^{-4}$ produces the best results---further supports this interpretation.

\subsection{Why Rank Distillation Cannot Stand Alone}

Rank distillation alone collapses entirely (Recall@10 $= 0.001$), despite being a well-motivated objective. This is because rank distillation operates on \emph{relative} document scores without anchoring the predicted embeddings to any absolute location in the embedding space. Without alignment loss to establish a reasonable initial embedding geometry, the rank distillation gradients provide no useful signal---the model cannot learn to rank documents correctly if its embeddings are not yet in a meaningful region of the space. However, once alignment establishes the geometric foundation, rank distillation provides complementary gains by refining the local ranking structure.

\section*{Limitations}

Our work has several limitations. First, we evaluate on a single dataset (QReCC); generalization to other conversational search benchmarks or open-domain retrieval tasks remains to be demonstrated. Second, our experiments use a same-family setting (Qwen3-8B and Qwen3-Embedding-8B), which may benefit from shared pretraining representations. Cross-family settings (e.g., projecting from one LLM family into an unrelated embedding space) are likely more challenging and are left for future work. Third, while the quality gap is small (--3.0\% to --3.6\%), it is statistically significant ($p = 0.0005$), meaning our method does not yet fully match the baseline. Fourth, the projection head requires training data generated by running both the LLM and the embedding model, so the embedding model is still needed during the training phase---only inference-time deployment is simplified. Finally, we have not explored whether the projection head can generalize to documents outside the training distribution without retraining.


\bibliography{custom}

\appendix

\section{Full Experiment Results}
\label{sec:full-results}

Table~\ref{tab:full-results} presents all 12 experimental configurations sorted by Recall@10, with bootstrap 95\% confidence intervals for the gap to the baseline.

\begin{table*}[t]
  \centering
  \begin{tabular}{llcccccc}
    \toprule
    \textbf{Group} & \textbf{Experiment} & \textbf{R@10} & \textbf{$\Delta$ R@10} & \textbf{95\% CI} & \textbf{MRR} & \textbf{$\Delta$ MRR} & \textbf{95\% CI} \\
    \midrule
    C & recipe\_e80\_lr2e4 & \textbf{0.607} & --0.030 & [--0.047, --0.014] & \textbf{0.293} & --0.036 & [--0.047, --0.024] \\
    A & All three losses & 0.595 & --0.042 & [--0.058, --0.025] & 0.288 & --0.040 & [--0.052, --0.029] \\
    B & Align-heavy RD & 0.591 & --0.046 & [--0.063, --0.029] & 0.280 & --0.049 & [--0.060, --0.037] \\
    B & RD top-256 & 0.584 & --0.053 & [--0.070, --0.037] & 0.284 & --0.045 & [--0.057, --0.033] \\
    A & Align + Contra & 0.582 & --0.055 & [--0.072, --0.038] & 0.281 & --0.047 & [--0.059, --0.036] \\
    A & Align + Rank & 0.579 & --0.058 & [--0.075, --0.041] & 0.272 & --0.057 & [--0.069, --0.045] \\
    A & Align only & 0.567 & --0.070 & [--0.088, --0.053] & 0.273 & --0.056 & [--0.068, --0.044] \\
    C & recipe\_e50 & 0.552 & --0.085 & [--0.102, --0.068] & 0.261 & --0.067 & [--0.080, --0.055] \\
    B & RD-heavy & 0.516 & --0.121 & [--0.140, --0.102] & 0.236 & --0.093 & [--0.106, --0.079] \\
    A & Contrastive only & 0.498 & --0.139 & [--0.159, --0.119] & 0.231 & --0.097 & [--0.110, --0.084] \\
    A & Rank distill only & 0.001 & \multicolumn{2}{c}{collapsed} & 0.000 & \multicolumn{2}{c}{collapsed} \\
    C & recipe\_lr5e4 & 0.001 & \multicolumn{2}{c}{collapsed} & 0.000 & \multicolumn{2}{c}{collapsed} \\
    \bottomrule
  \end{tabular}
  \caption{Full results for all 12 configurations across three experiment groups. Group A: loss ablation (30 epochs, lr $3 \times 10^{-4}$). Group B: loss weight tuning (30 epochs, lr $3 \times 10^{-4}$, $\lambda_c = 0$). Group C: training recipe variations. Baseline: Recall@10 = 0.637, MRR@10 = 0.329.}
  \label{tab:full-results}
\end{table*}

\section{Implementation Details}
\label{sec:impl-details}

\paragraph{Hidden State Extraction.}
The LLM generates queries via greedy decoding with a maximum of 32 new tokens. At each autoregressive step, we extract the last-layer hidden state ($d_h = 4096$) and retain only non-special tokens (excluding BOS, EOS, and padding). KV caching is used for efficient generation. All hidden states are stored as float32.

\paragraph{Trace Caching.}
To enable rapid experimentation, we cache the LLM's generation traces (hidden states, query text, token counts) as per-trigger \texttt{.npz} files, keyed by a hash of the model name, tokenizer max length, and generation length. A total of 11,017 traces are cached. On cache hit, the LLM is not loaded at all, reducing experiment time from $\sim$2 hours to $\sim$15--20 minutes.

\paragraph{Projection Head.}
The transformer encoder uses pre-norm layer normalization, 8 attention heads, and a feedforward dimension of $4 \times d_m = 4096$. Dropout is set to 0. Learnable positional embeddings (initialized to zeros) support sequences up to 128 tokens. The output projection maps from $d_m = 1024$ to $d = 1024$ (the embedding model's output dimension), followed by L2 normalization.

\paragraph{Training.}
We use AdamW with $\beta_1 = 0.9$, $\beta_2 = 0.999$, weight decay $10^{-4}$, and gradient clipping at norm 1.0. The learning rate follows a cosine schedule from $2 \times 10^{-4}$ to $10^{-5}$ over 80 epochs. Batch size is 16. For rank distillation, the top-$K = 128$ documents are retrieved per query using the teacher's embeddings, and their similarity scores serve as the soft targets.

\paragraph{Evaluation.}
Bootstrap confidence intervals use 1,000 resamples with replacement at the trigger level. McNemar's test compares per-trigger binary retrieval success (whether the relevant document appears in the top-10 results) between our method and the baseline. All experiments run on a single NVIDIA RTX PRO 6000 (Blackwell architecture, 98GB VRAM).

\end{document}